\documentclass[10pt, a4paper]{article}

\usepackage[final]{lrec2026} 

\usepackage[inline]{enumitem} 
\usepackage{amssymb}
\usepackage{multirow}
\usepackage{booktabs} 
\usepackage{graphicx}      
\newcolumntype{L}[1]{>{\raggedright\arraybackslash}p{#1}}
\usepackage{newunicodechar}
\usepackage{textcomp}
\usepackage{underscore} 
\newunicodechar{≤}{\leq}
\newunicodechar{≈}{\approx}

\title{LuxBorrow: From Pompier to Pompjee, Tracing Borrowing in Luxembourgish}

\name{
  Nina Hosseini-Kivanani$^{* 1,2}$\thanks{* Both authors contributed equally to this work.},
  Fred Philippy$^{*3}$\footnotemark[1]
}

\address{$^{1}$ University of Luxembourg, Luxembourg\\
         $^{2}$ Radio Télévision Luxembourg (RTL), Luxembourg\\
         $^{3}$ SnT, University of Luxembourg, Luxembourg\\
         {\small nina.hosseinikivanani@ext.uni.lu, fred.philippy@uni.lu}}
\abstract{We present \textsc{LuxBorrow}, a borrowing-first analysis of Luxembourgish (LU) news spanning 27 years (1999--2025): 259{,}305 RTL articles and 43.7M tokens. Our pipeline combines sentence-level language identification (LU/DE/FR/EN) with a token-level borrowing resolver restricted to LU sentences, using lemmatization, a collected loanword registry, and compiled morphological/orthographic rules. Empirically, LU remains the matrix language across all documents, while multilingual practice is pervasive: 77.1\% of articles include at least one donor language and 65.4\% use three or four. Breadth does not imply intensity: median code-mixing index (CMI) increases from 3.90 (LU+1) to only 7.00 (LU+3), indicating localized insertions rather than balanced bilingual text. Domain/period summaries show moderate but persistent mixing, with CMI rising from 6.1 (1999--2007) to a peak of 8.4 (2020). Token-level adaptations total 25{,}444 instances and exhibit a mixed profile: morphological 63.8\%, orthographic 35.9\%, lexical 0.3\%; the most frequent single rules are orthographic (\textit{on→oun}, \textit{eur→er}), while morphology is collectively dominant. Diachronically, code-switching intensifies, and morphologically adapted borrowings grow from a small base; French overwhelmingly supplies adapted items, with modest growth for German and negligible English. We advocate borrowing-centric evaluation, borrowed token/type rates, donor entropy over borrowed items, and assimilation ratios over headline document-level mixing indices.
\newline 
}

\begin{document}

\maketitleabstract

\section{Introduction}
Lexical borrowing, the adoption of lexical items from a donor language, often with orthographic or morphological adaptation, is pervasive in multilingual media and directly affects the robustness of downstream NLP systems. In the Luxembourgish (LU) context, where sustained contact with German (DE) and French (FR) structures everyday communication in administration, education, and the press, borrowing is common yet insufficiently studied at scale. While code-switching co-occurs in this ecology, we foreground borrowing as our primary object of analysis and treat code-switching descriptively to contextualize borrowing rates. This choice follows the ``Simple View'' of borrowing and code-switching, which emphasizes that borrowings are typically lexically listed and grammatically integrated, whereas code-switches are more spontaneous and syntactically independent~\cite{treffers2025simple}. Prior work in multilingual European communities, including studies of Portuguese-speaking minorities in Luxembourg, further indicates that insertional code-switching patterns vary with language dominance, underscoring the need to distinguish code-switching from entrenched borrowings~\cite{stell2012code}. More broadly, European multilingualism often blurs this boundary as speakers integrate foreign items in ways that challenge categorical distinctions~\cite{chan2025borrowing}.

We take a borrowing-first view of multilingual LU news and contribute the following:
\begin{enumerate*}[label=\roman*), itemjoin={{, }}, itemjoin*={{, and }}]
\item we annotate a 27-year LU news corpus (1999–2025) with hybrid token-level Language identification (LID) (LU/DE/FR/EN) and explicit borrowing/code-switching labels;
\item we implement a morphological pattern matching system that distinguishes borrowed words from genuine code-switches using contextual analysis (foreign run length, LU neighborhood ratios);
\item we present comprehensive code-switching metrics (Code-Mixing Index (CMI), entropy, M-index) with temporal evolution analysis; and
\item we develop a hybrid language identification approach combining sentence-level LID with token-level refinement and lexicon-based fallback for multilingual news text processing.


\end{enumerate*}

We organize our investigation around the following research questions:

\begin{itemize}

\item RQ1: How frequent is lexical borrowing across news domains and time periods, and how does borrowing integration compare to general code-switching patterns in multilingual contexts?
\item RQ2: What are the patterns of language mixing at the document level in LU news texts, and to what extent does LU remain the dominant matrix language?
\item RQ3: Can morphological adaptation patterns effectively distinguish established borrowing from spontaneous code-switching in closely related Germanic languages (LU/DE)?
\item RQ4: How do borrowing integration patterns evolve diachronically (1999–2025), and what competitive dynamics emerge between borrowed forms and native LU alternatives?
\end{itemize}


Prior research on borrowing and code-switching spans descriptive linguistics, sociolinguistics, and computational modeling. Foundational accounts distinguish entrenched borrowings from in-situ alternation, viewing the two as a continuum governed by lexical integration and community norms~\cite{treffers2025simple,masojc2023borrowings,deuchar2020code}. Recent NLP studies operationalize these phenomena through token-level language identification, normalization, and cross-lingual robustness benchmarks, often blurring linguistic boundaries for pragmatic classification~\cite{winata2023decades,prabhugaonkar2017differentiating}. We build on both strands, with emphasis on resources and methods applicable to low-resource settings where code-switching is prevalent and linguistic resources are scarce.

\section{Related Work}
\paragraph{Borrowing vs. code-switching.}
Contact linguistics distinguishes \emph{lexical borrowing}, items integrated into the recipient language’s lexicon and grammar, from \emph{code-switching}, i.e., spontaneous alternation between languages within discourse. Foundational work~\cite[e.g.,][]{poplack2017borrowing,poplack1988social,poplack1984borrowing} shows that entrenched borrowings exhibit morphological/phonological integration and community-wide diffusion, whereas code-switches retain structural independence and are often speaker-specific. The ``Simple View'' further operationalizes this distinction via \emph{listedness} (membership in the mental lexicon)~\cite{treffers2025simple}. Concretely, lexically listed and community-entrenched items, attested in dictionaries, frequent, and morphologically/phonologically integrated into the recipient language, are classified as borrowings, whereas unlisted, ad hoc insertions that retain donor-language structure are treated as code-switches.

\paragraph{From generic mixing to borrowing identification.}
NLP has shifted from global ``mixing'' indices to explicit borrowing detection. \citet{alvarez2022detecting} introduce Spanish newswire corpora annotated for unassimilated borrowings (``anglicisms'') and benchmark CRF, BiLSTM-CRF, and Transformer taggers. This line seeded the ADoBo shared task at IberLEF and released public mBERT taggers~\cite{de2021adobo}. Beyond monolingual scenarios, low-resource loanword discovery uses phonetic/semantic similarity to propose candidates with minimal supervision~\cite{miller2021neural,mi2020loanword}. In parallel, work on social-media corpora distinguishes borrowings from code-switches and integrates borrowing signals into sequence labeling pipelines~\cite{kent2018incorporating}. Together, these efforts establish borrowing detection as a distinct token-level task rather than a by-product of LID.

Code-switching research popularized document-level indices (e.g., CMI), but subsequent analyses highlight their sensitivity to utterance length and annotation noise and their weak interpretability for edited text~\cite{srivastava2021challenges,thara2018code,chandu2018code}. Borrowing-centric studies instead report \emph{itemized} statistics: the proportion of borrowed tokens and types, donor-language composition, and evidence of assimilation via morphological integration~\cite{poplack1988social}. Following this tradition, we report \begin{enumerate*}[label=\roman*), itemjoin={{, }}, itemjoin*={{, and }}] \item Borrowed Token Rate and Borrowed Type Rate \item donor-language entropy over borrowed items~\cite{rosillo2025entropy} \item an \emph{assimilation ratio}, the share of borrowed tokens showing lexical/morphological integration. \end{enumerate*}

Because borrowing identification often requires donor-language labels and disambiguation of orthographic homographs, we treat token-level LID as a \emph{supporting} component rather than the primary objective, aligning with prior distinctions between code-switching focused LID and granular borrowing detection~\cite{treffers2025simple,kent2018incorporating}. Standard code-switiching benchmarks (LinCE, GLUECoS) remain useful for pretraining signals and protocol alignment (LID, Part-of-Speech (POS), Named Entity Recognition (NER)), even if their targets are broader than borrowing per se~\cite{khanuja2020gluecos,aguilar2020lince}. Extensions to joint LID+POS~\cite{soto2018joint} and subword/intra-word LID~\cite{sabty2021language,mager2019subword} further improve robustness in morphologically complex settings~\cite{burchell2024code}.

For LU, resources have recently expanded. LuxBank provides the first UD treebank for syntax-aware modeling~\cite{plum2024luxbank}. The Lëtzebuerger Online Dictionnaire (LOD) supplies lexicon attestations relevant to entrenchment analyses~\citelanguageresource{LOD}. Neural text normalization (TN) addresses LU’s orthographic variation and improves downstream robustness for LID and borrowing detection~\cite{lutgen2025neural}. In generation tasks (e.g., LU headlines), numeric/date/currency TN quality matters for factual consistency; recent work emphasizes explicit factuality checks (e.g., FRANK/AlignScore) and decoding strategies that reduce hallucination~\cite{malon2024self,zha2023alignscore,plum2025text}.

\paragraph{Gap and our contribution.}

Current code-switching benchmarks and the ADoBo framework do not close the borrowing-first gap: resources are sparse, and no borrowing-labeled corpus exists for LU~\cite{alvarez2022detecting,de2021adobo,mellado2022borrowing,mi2020loanword}. We try to fill this gap with a token-level, borrowing-centric LU corpus and modeling baselines, using LOD to enable reproducible diachronic analysis and cross-linguistic comparisons beyond well-resourced pairs such as Spanish–English~\cite{list2022automated}.

\begin{figure*}[!htp]
\begin{center}
\includegraphics[width=0.9\textwidth]{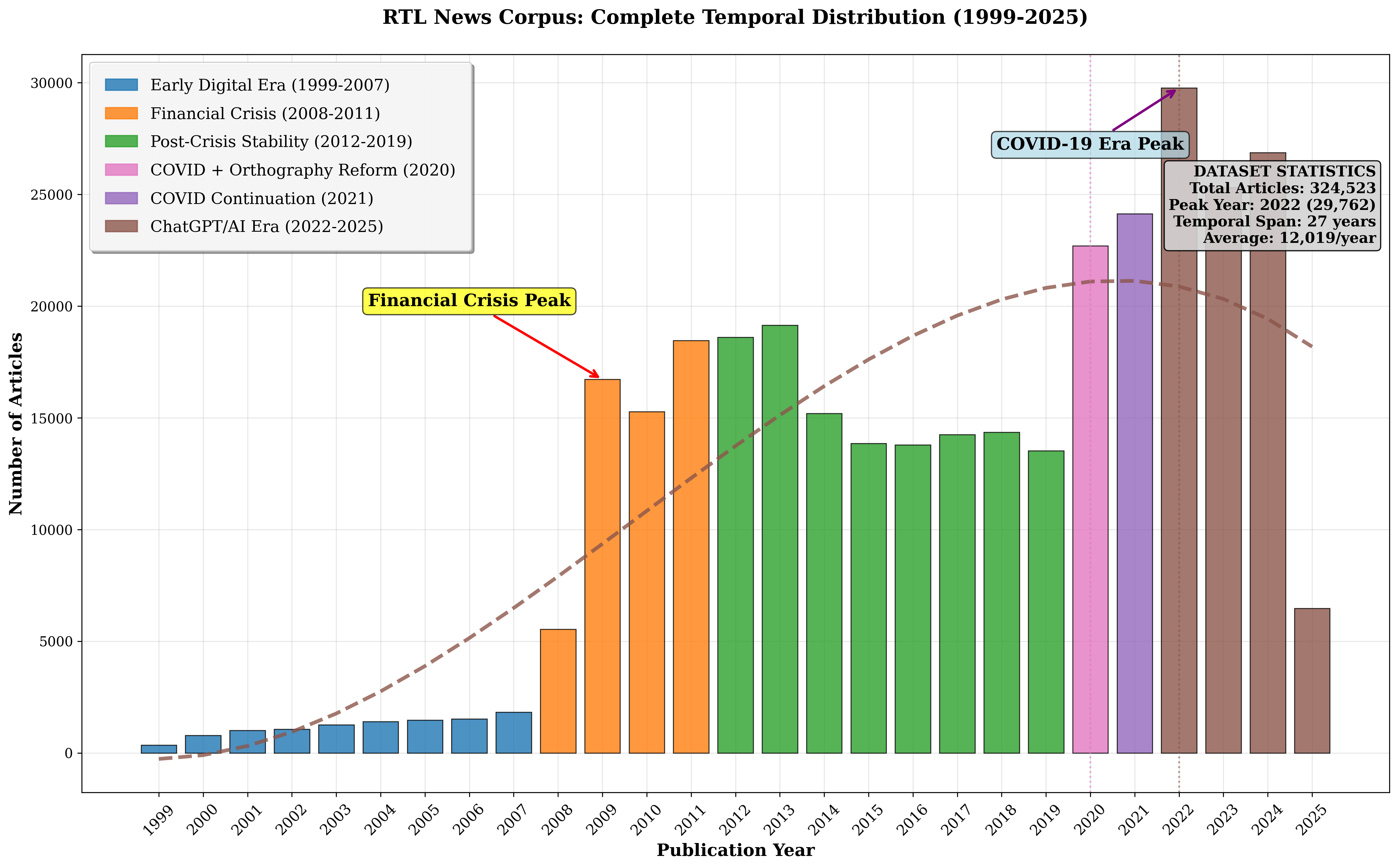}
\caption{Temporal distribution of RTL news articles (1999--2025) with selected linguistic and technological milestones. The brown dotted line marks the 2020 orthography reform and onset of the COVID 19 period.}

\label{fig.1}
\end{center}
\end{figure*}

\section{Materials \& Methodology}
We study multilingual mixing in LU news by: \begin{enumerate*}[label=\roman*), itemjoin={{, }}, itemjoin*={{, and }}] \item identifying language at sentence and token level \item operationalizing lexical borrowing versus code-switching \item quantifying diachronic and domain-specific trends.\end{enumerate*} We treat \emph{borrowing} as lexically integrated items (orthographic/morphological accommodation to LU or entrenched lexicalized forms) and \emph{code-switching} as spontaneous alternation without LU-internal integration.

\subsection{Dataset}


This study uses a large-scale longitudinal corpus of RTL.lu news articles (1999--2025). The corpus was \emph{not} constructed to elicit multilingual behavior; rather, as professionally edited LU reporting produced in Luxembourg’s trilingual media environment, it naturally contains instances of lexical borrowing and occasional code-switching. We therefore treat it as an observational resource for contact-linguistic phenomena in contemporary written LU. The RTL news corpus comprises 259,305 articles distributed across six distinct socio-historical periods: Early Digital Era (1999-2007), Financial Crisis period (2008-2011), Post-Crisis Stability (2012-2019), COVID-19 pandemic coinciding with LU orthography reform (2020), COVID continuation (2021), and the ChatGPT/AI Era (2022-2025). This temporal stratification enables diachronic analysis of contact linguistic phenomena across major socio-economic and technological transitions, with notable publication peaks during the 2009 Financial Crisis (16,713 articles) and the 2022 COVID-era surge (29,762 articles). Figure~\ref{fig.1} shows the uneven article volume over time and highlights key milestones such as the 2020 orthography reform (brown dotted line).



\subsection{Language Identification}
We adopt a hierarchical two-stage pipeline: (1) sentence-level LID as a context gate, and (2) token-level borrowing detection inside LU sentences.
\begin{enumerate}[label=\arabic*)]
    \item {Sentence-level LID (context gate).}
We use OpenLID (FastText-based) to predict the sentence language \citep{burchell-etal-2023-open}. To curb LU over-assignment on short sentences, we apply a length-adaptive posterior threshold: a base of 0.50 is increased up to 0.80 for short inputs; low-confidence cases are routed to Other. Only sentences labeled LU by this gate proceed to the token-level detector.

\item {Token-level borrowing detection (within LU sentences).}
Rather than generic token LID, we run a borrowing detector on LU sentences. Tokens are lemmatized and morphologically normalized with \textit{spellux} \citep{spellux2020}, then matched against a curated lexicon of 7\,796 loanword entries with donor tags (FR/DE/EN) and assimilation patterns. Each token receives one of \textsc{Native}, \textsc{FR\_LOAN}, \textsc{DE\_LOAN}, or \textsc{EN\_LOAN}. The lexicon is versioned and frozen for this study.

\end{enumerate}


To distinguish between entrenched lexical borrowings and code-switching events, we apply a compiled set of morphological and lexical patterns derived from empirical observations in the corpus. For each candidate token, the system computes contextual features including the run-length of foreign-language tokens and the local LU token ratio within a ±3-token window. Tokens matching a known borrowing pattern and occurring in predominantly LU contexts (short foreign spans and high local LU density) are classified as borrowings. Longer foreign spans are marked as code-switching, while inconclusive cases remain ambiguous.

Concretely, we treat items such as \textit{De Sträit ass duerch e \underline{Malentendu} entstan.} (\textbf{EN}: \textit{The fight occurred due to a \underline{misunderstanding}.}), as containing an entrenched borrowing, where \textit{Malentendu} behaves like a LU noun in an otherwise LU sentence.
By contrast, a sentence such as \textit{D'Buch, \underline{ça n'a rien à voir} mat dem Film.} (\textbf{EN}: \textit{The book \underline{has nothing to do} with the movie.}) is analyzed as a French code-switching span because it forms a longer contiguous FR clause with donor language syntax.

\subsection{Loanword Identification}
\paragraph{Seed Data}
Currently, there is no official list of loanwords in LU, even though it is widely acknowledged that the language contains a substantial number of borrowings from other languages. To identify such loanwords, we relied on the LOD 
as our primary resource. The LOD provides LU entries together with their possible multiple meanings and translations into German, French, English, and Portuguese. Since loanword borrowing into LU predominantly occurs from DE, FR, and EN, we restricted our analysis to these three languages and disregarded Portuguese. We further limit the scope to nouns, verbs, and adjectives, as these parts of speech are the most frequent categories for lexical borrowing. Dictionary entries explicitly marked as proper nouns (e.g., chemical elements, brand names, currencies, places, measurement units, etc.) are excluded as well.

\paragraph{Morphological and Orthographic Adaptation Patterns}
Loanwords in LU are either taken over directly from the donor language or frequently undergo systematic morphological or orthographic adaptations when integrated from source languages (e.g., FR \textit{ajuster} $\rightarrow$ LU \textit{ajustéieren} via the \textit{-er $\rightarrow$ -éieren} adaptation). 
For each LU entry, we examine its translations into DE, FR, and EN to determine whether the relationship corresponds to one of these adaptation patterns. 

\paragraph{Parallel Borrowing}
In some cases, a candidate match occurs simultaneously with more than one language. Such situations arise mainly for two reasons: (1) they often involve words that were independently derived across languages from a common root (e.g., \textit{talentéiert}, \textit{talentiert} and \textit{talented} are borrowed from Latin \textit{talentum} and Ancient Greek \textit{tálenton}), or from another source language (e.g., \textit{Alibi}, \textit{Anorak}, or \textit{Tsunami}). These cases do not represent genuine instances of borrowing into LU; (2) they reflect secondary borrowing chains. For example, the LU word \textit{Successioun} matches with both FR and EN \textit{succession}, but the EN form is itself a borrowing from FR.

Without accounting for such cases, these words would misleadingly appear as parallel borrowings rather than revealing the true donor-recipient relationship. To address this, we compiled external lists of known loanwords across the three languages \citelanguageresource{Wiktionary_EN_from_FR, Wiktionary_FR_from_EN, Wiktionary_DE_from_FR, Wiktionary_DE_from_EN}, covering the most common borrowing directions: EN~$\rightarrow$~DE, FR~$\rightarrow$~DE, EN~$\rightarrow$~FR, and FR~$\rightarrow$~EN. This step is crucial not only for identifying which words are loanwords, but also for determining from which language they were originally borrowed.

\paragraph{Shared Inheritance}
Another issue that must be addressed concerns the linguistic proximity between LU and DE, which stems from their shared West Germanic origin. As a result, many LU words are identical or nearly identical to their DE equivalents. This makes it difficult to determine whether a given term is genuinely a loanword or whether it developed independently in both languages from a common Old High German root, without any borrowing taking place. To resolve this, words previously identified as DE loanwords are reclassified as non-loanwords if the corresponding DE term can be traced back to Old High German~\citelanguageresource{Wiktionary_Old_High_German}, since it is likely that the LU form also evolved directly from the same ancestral source rather than being borrowed from modern DE.

\paragraph{Human Annotation}
After the automatic loanword detection, an initial identification and frequency count of the detected words within the corpus is carried out. A human annotator then reviewed these results to identify major shortcomings. The most significant issues included: (1) relevant and frequently occurring loanwords that were not captured by the automatic pattern-matching procedure (e.g., \textit{Pompjee} from FR \textit{pompier}, \textit{Grupp} from FR  \textit{groupe}); (2) frequent compound words absent from the seed dictionary data (e.g., \textit{Policepatrull} from FR \textit{police} and \textit{patrouille}); and (3) common orthographic variants or misspellings that, despite being nonstandard, occur frequently (e.g., \textit{entretemps} instead of \textit{entre-temps}, \textit{Akteur} instead of \textit{Acteur}).
The annotator also removed words incorrectly tagged as loanwords by the automatic pipeline.

\paragraph{Final Corpus of Loanwords}
Using this strategy, we identify approximately 3,632 loanwords of DE origin, about 3,201 from FR, and around 535 from EN. For each tagged word, its officially documented spelling variants are also extracted from the source dictionary.

\begin{figure*}[ht!]
    \centering
    \includegraphics[width=0.77\linewidth]{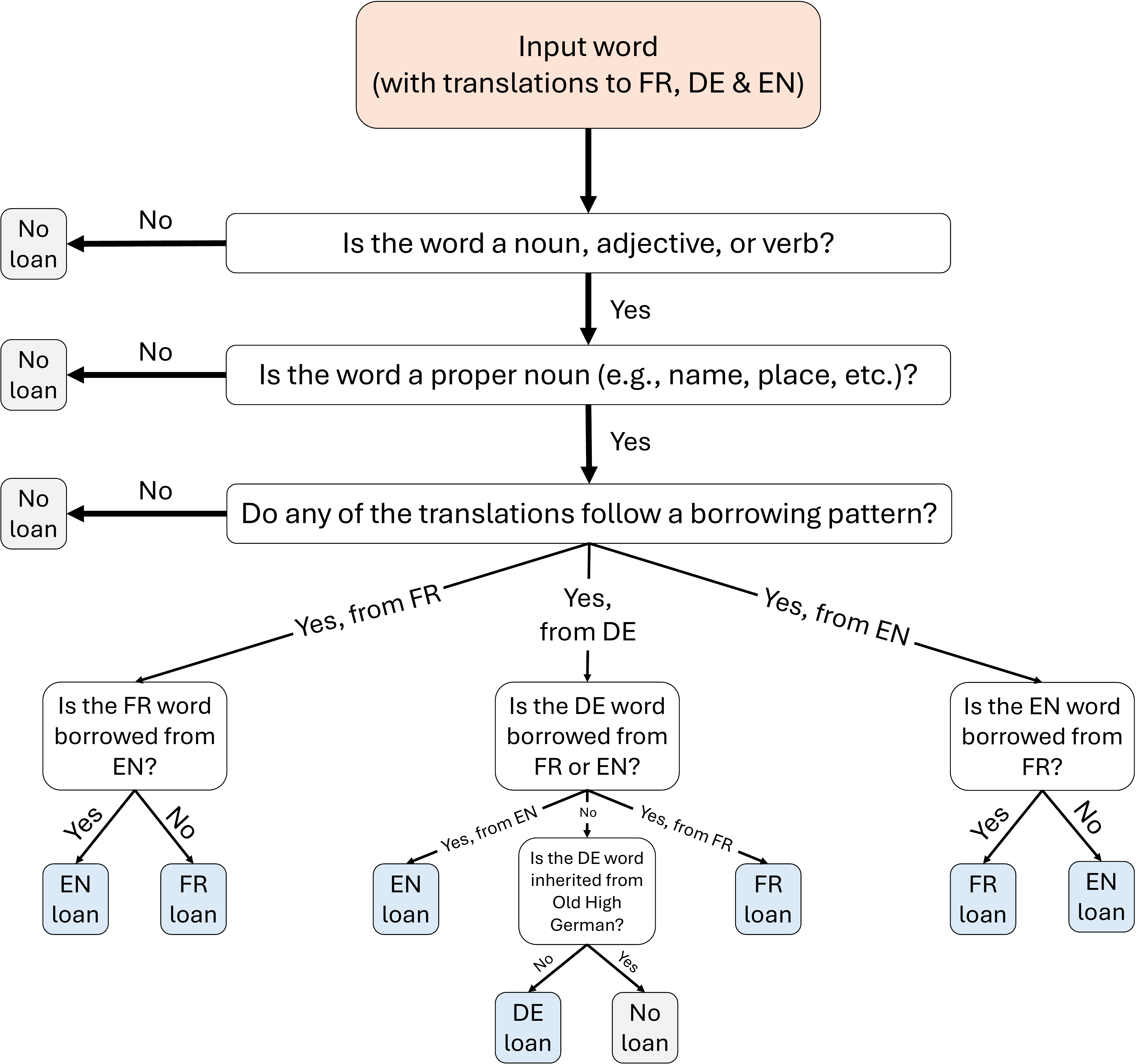}
    \caption{Pipeline for the initial automatic loanword detection. After completing this stage, the resulting loanword corpus was manually reviewed to identify major shortcomings, such as incorrectly tagged words, missing loanwords, or incomplete compound forms, and adjusted accordingly.}
    \label{fig:loanword_detection_pipelin}
\end{figure*}

\subsection{Statistical analysis}
Our analysis combines descriptive statistics with selective inferential methods and advanced indexing. We implement compiled pattern indexing for constant-time morphological lookup and bidirectional lexical indexing for efficient retrieval across 43\,M+ tokens. In total, we process 259{,}305 RTL articles comprising 43.7\,M tokens. Token-level language distribution (OpenLID): LU 34.58\,M (92.5\%), FR 1.41\,M (3.8\%), DE 1.08\,M (2.9\%), and EN 0.31\,M (0.8\%). For RQ1 and RQ4, we compute document- and section-level code-mixing metrics, CMI, Shannon entropy, and M-Index, then aggregate them by news section and predefined year buckets (1999–2007, 2008–2011, 2012–2019, 2020, 2021, 2022–2025), exporting per-group means and token/article counts. For RQ2, we perform document-level code-mixing analysis using OpenLID token-level language labels (LU, DE, FR, EN). We calculate CMI, entropy, and M-index for each document and analyze multilingual scope patterns by comparing monolingual vs. multilingual documents. The analysis includes language combination frequencies, donor language prevalence, and mixing intensity distributions across the corpus using descriptive statistics. For RQ3, we tabulate borrowing frequencies by donor language (FR/DE/EN) and list the most frequent borrowed forms and associated morphological patterns. We compute several established code-switching metrics, including the CMI, Shannon entropy, and the M-Index. These are aggregated by domain (e.g., politics, sports, culture) and by temporal buckets: 1999–2007, 2008–2011, 2012–2019, 2020, 2021, and 2022–2025. Additionally, monthly aggregates are computed for diachronic tracking of code-switching rates, borrowing shares, and donor language distributions. Indexing systems enable real-time pattern matching and scalable lexical analysis, while outputs are summary tables and statistical comparisons intended for downstream visualization and hypothesis testing.

\section{Results \& Discussion}
Multilingual mixing in LU news is measurable and domain-sensitive, but sensitive to small-sample spikes. We therefore report summaries for overall mixing (RQ1/RQ4), document-level patterns (RQ2), and borrowing composition/assimilation (RQ3), noting where estimates are driven by sparse events.

\paragraph{RQ1:} 

The analysis of 259,305 RTL news articles (1999–2025) reveals clear domain-level contrasts in multilingual mixing.
Political subdomains exhibit the strongest code-mixing effects: CSV Süden reaches the highest which 38.5, PK Juncker shows the highest entropy (0.882), and DP Süden 2009 achieves the most balanced distribution (M=0.167). However, these extreme values are the result of small-sample events ($n\leq 3$), where a single multilingual speech or press release can inflate the metrics.

In contrast, large-scale sections such as National (52K articles, CMI=8.8) and International (45K articles, CMI=8.0) show stable yet persistent mixing across decades. Temporarily, the intensity of code mixing has increased from CMI=6.1 in 1999–2007 to a peak of 8.4 in 2020 (+38\%), with corresponding rises in entropy (0.269 → 0.336) and M-index (0.022 → 0.035), indicating greater lexical diversity and more balanced multilingualism in recent years.


Overall, political coverage dominates multilingual expression, while national and international sections maintain moderate but consistent integration of lexical material from FR, DE, and EN. Table~\ref{tab:rq1_table} summarizes CMI, entropy, and M-index values for representative domains and periods.

\begin{table}[th!]
\centering
\footnotesize
\resizebox{\columnwidth}{!}{%
\begin{tabular}{@{}lrrrr@{}}
\toprule
\textbf{Domain / Period} & \textbf{CMI} & \textbf{Entropy} & \textbf{M-Index} & \textbf{\#Articles} \\
\midrule
CSV Süden (Politics)     & \textbf{38.46} & 0.859 & 0.1542 & 1 \\
PK Juncker(Press Conf.) & 34.62          & \textbf{0.882} & 0.077 & 1 \\
DP Süden(Politics)      & 16.67          & 0.451 & \textbf{0.167} & 2 \\
National                 & 8.82          & 0.351 & 0.035 & \textbf{52,376} \\
International            & 8.00          & 0.322 & 0.035 & 45,486 \\
\addlinespace
\textbf{1999--2007}      & 6.10          & 0.269 & 0.022 & 9,795 \\
\textbf{2020 (peak)}     & \textbf{8.40} & 0.336 & 0.035 & 17,831 \\
\bottomrule
\end{tabular}%
}
\caption{Computed code-mixing intensity and diversity (CMI, entropy, M-Index) for RTL news articles by domain and temporal slice.}
\label{tab:rq1_table}
\end{table}

\paragraph{RQ2:}
LU remains the matrix language across the corpus (100\% of documents). Multilingual practices are nevertheless pervasive: 77.1\% of documents mix LU with at least one of DE, FR, or EN, and 65.4\% use three or four languages within a single article. Despite this breadth, the intensity of mixing is low on average ($\mathrm{CMI}=5.25$), indicating LU dominance with localized insertions from donor languages rather than balanced bilingual text. 

\begin{table}[t]
\centering
\scriptsize
\renewcommand{\arraystretch}{1.01}
\setlength{\tabcolsep}{3.5pt}
\begin{tabular}{@{}lrrcc@{}}
\toprule
\textbf{Scope / Combination} & \textbf{Count} & \textbf{\%} & \textbf{Med.\ CMI} & \textbf{IQR} \\
\midrule
\multicolumn{5}{@{}l}{\textit{Document-level scope}} \\
LU only   & 74{,}946  & 22.9 & \textbf{0.00} & 0.00--0.00 \\
LU + 1    & 38{,}237  & 11.7 & 3.90 & 2.56--5.71 \\
LU + 2    & 109{,}367 & \textbf{33.5} & 6.67 & 4.84--8.82 \\
LU + 3    & 104{,}189 & \textbf{31.9} & \textbf{7.00} & 5.41--8.90 \\
\addlinespace[2pt]
\multicolumn{5}{@{}l}{\textit{Most frequent donor combinations}} \\
FR+LU        & 21{,}830 & 8.7  & -- & -- \\
DE+LU        & 13{,}496 & 5.4  & -- & -- \\
EN+LU        & 2{,}911  & 1.2  & -- & -- \\
DE+FR+LU     & 93{,}138 & \textbf{37.0} & -- & -- \\
EN+FR+LU     & 10{,}268 & 4.1  & -- & -- \\
DE+EN+FR+LU  & 104{,}189 & \textbf{41.4} & -- & -- \\
\bottomrule
\end{tabular}
\vspace{3pt}
\caption{Document-level language mixing and donor composition. LU dominates even in multilingual articles. FR and DE co-occur most frequently, with EN as a secondary donor.}
\label{tab:rq2}
\end{table}

CMI rises monotonically with the number of co-present languages, but remains low in absolute terms even in highly multilingual documents. LU-only articles are, by definition, unmixed ($\mathrm{CMI}=0$). Median CMI increases from 3.90 in LU+1 to 6.67 in LU+2 and 7.00 in LU+3 (Table~\ref{tab:rq2}). This shows that \emph{breadth} (three–four languages) does not entail \emph{high intensity} of mixing: LU continues to dominate, with insertions from donor languages rather than balanced bilingual text.

Among multilingual documents ($n=251{,}793$), FR is the most frequent donor language (91.1\%), followed by DE (86.1\%) and EN (49.0\%). At the combination level, the fully multilingual set (LU+DE+FR+EN) accounts for 41.4\% of multilingual documents, while the most common triad is DE+FR+LU (37.0\%). Dyads are comparatively rare (15.2\%), led by FR+LU (8.7\%) and DE+LU (5.4\%). Overall, FR and DE jointly drive most document-level multilinguality, with EN as a secondary co-present donor.

\begin{figure}[hpt!]
    \centering
    \includegraphics[width=1.01\linewidth]{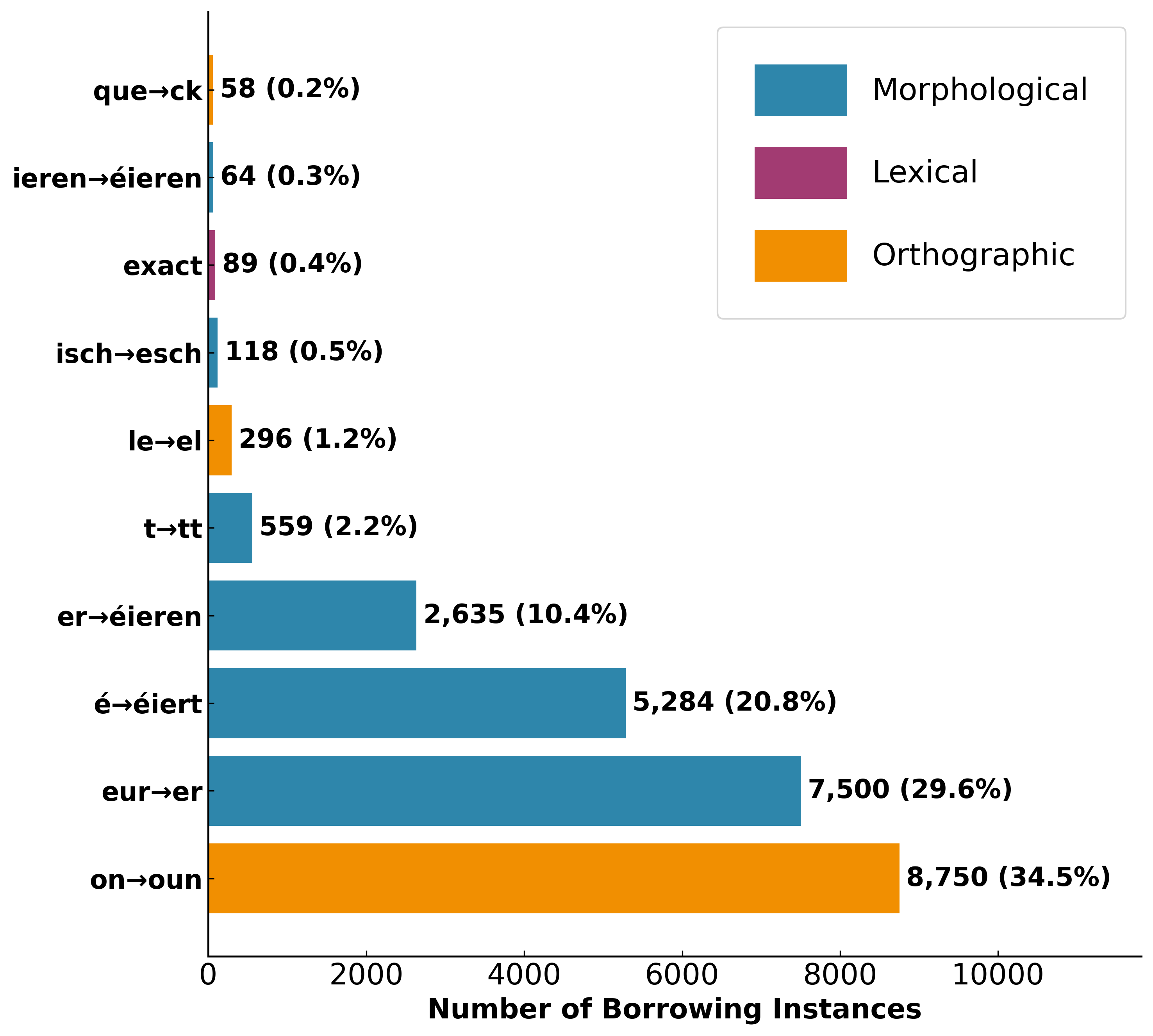}
    \caption{Distribution of observed borrowing pattern types across the RTL news corpus. Borrowing adaptation types across.}
    \label{fig:rq3_borrowing_patterns}
\end{figure}

\paragraph{RQ3:}
Using our token-independent, rule-based detector on LU sentences, we identified 25{,}444 borrowing instances distributed across three adaptation types: morphological (16{,}221; 63.8\%), orthographic (9{,}134; 35.9\%), and lexical (89; 0.3\%). Within the \emph{ten most frequent concrete patterns} (\(n=25{,}353\)), the distribution shows a mixed morphological-orthographic dominance: \textit{on→oun} (8{,}750; 34.5\%), \textit{eur→er} (7{,}500; 29.6\%), and \textit{é→éiert} (5{,}284; 20.8\%) together account for \(\approx 85\%\) of the top-10 set. Orthographic shifts such as \textit{on→oun} (8{,}750; 34.5\%), \textit{le→el} (296; 1.2\%), \textit{que→ck} (58; 0.2\%), and \textit{É→E} (30; 0.1\%) represent a substantial component of the adaptation inventory. Morphological patterns including \textit{er→éieren} (2{,}635; 10.4\%), \textit{t→tt} (559; 2.2\%), and \textit{isch→esch} (118; 0.5\%) remain productive but secondary. Pure lexical borrowings (unadapted forms) are minimal at both class level (0.3\%) and individual pattern frequency (\textit{exact}: 89; 0.4\%). Overall, the distribution in Figure~\ref{fig:rq3_borrowing_patterns} indicates a mixed orthographic–morphological profile: orthographic adaptation rules produce the highest individual counts, morphological patterns dominate collectively, and lexical borrowings form a minimal tail.

\begin{figure}[hpt!]
    \centering
    \includegraphics[width=\linewidth]{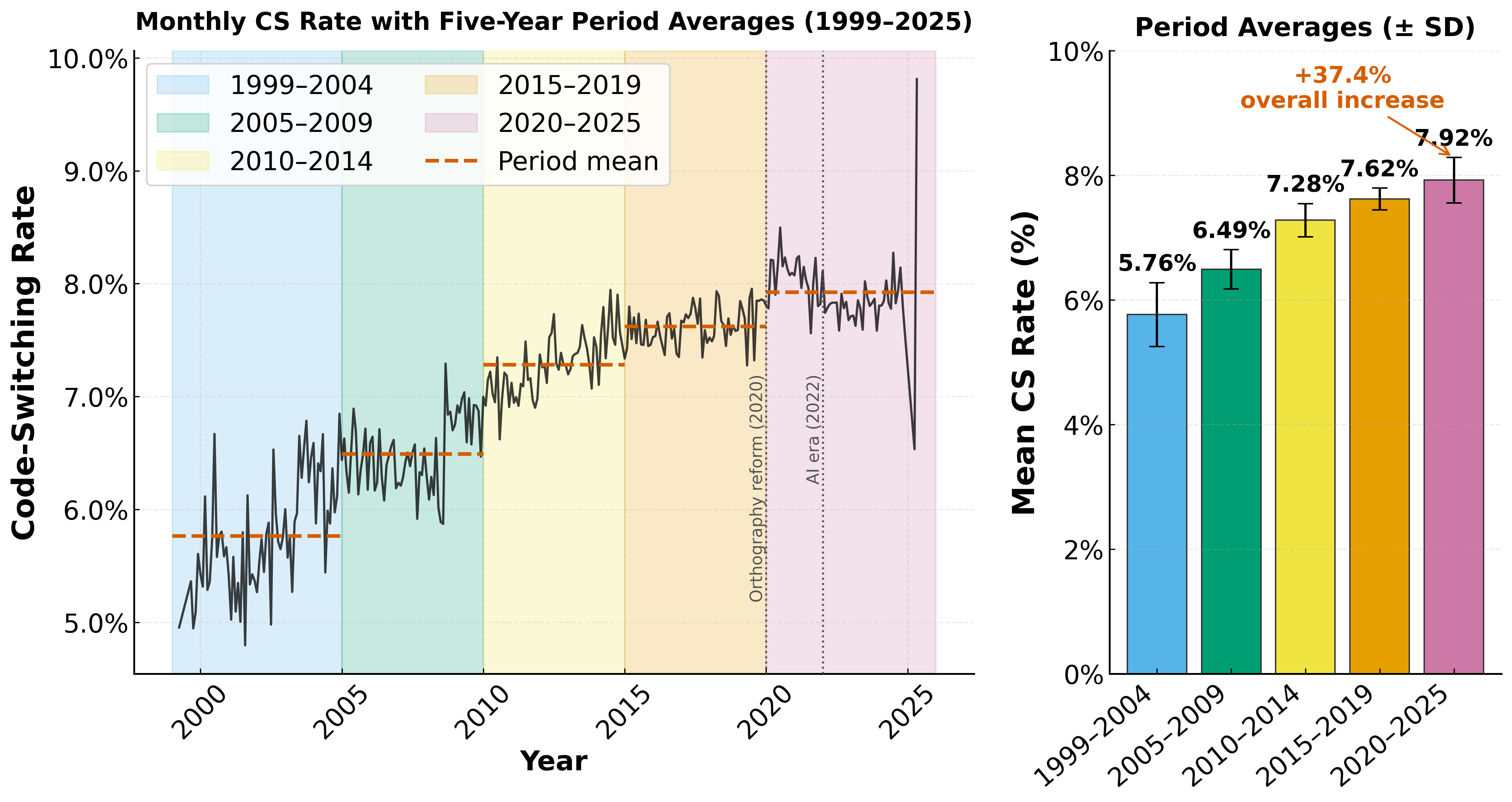}
    \caption{Diachronic evolution of code-switching in Luxembourgish news media (1999--2025). \textit{Left:} monthly CS rate
           with five-year period means (dashed lines). \textit{Right:} mean CS rate per
           period; error bars denote one standard deviation.
    \label{fig:rq4_period_trend}}
\end{figure}

\paragraph{RQ4:}
Figure~\ref{fig:rq4_period_trend} illustrates the diachronic evolution of code-switching in Luxembourgish news media across 307 months of data (1999--2025). The left panel displays the monthly CS rate (black line) with five-year analytical periods highlighted in distinct colors and period means overlaid as dashed orange lines; dotted vertical lines mark two external milestones, namely the 2020 orthography reform and the onset of the AI era in 2022. The right panel summarizes the mean CS rate and standard deviation for each interval. Code-switching metrics, calculated in five-year analytical intervals, demonstrate a consistent upward trend: 5.76\% (1999–2004), 6.49\% (2005–2009), 7.28\% (2010–2014), 7.62\% (2015–2019), and 7.92\% (2020–2025). This progression represents a 37.4\% increase in CS frequency over the 26 year period in this corpus, suggesting a gradual intensification of multilingual mixing in written discourse. Standard deviations across periods (ranging from 0.5114 in the early years to 0.1752–0.3642 more recently) suggest increasingly stable mixing patterns in most periods.

Morphological borrowing accounts for a small but measurable fraction of multilingual mixing. The borrowing share averages 0.73\% across all periods (range: 0.00–6.25\%), indicating that the vast majority of observed multilingual elements are insertional code-switches rather than morphologically adapted borrowings. This low borrowing share confirms that code-switching in LU news is predominantly characterized by direct lexical insertion rather than systematic morphological integration.

Donor language patterns in morphological borrowing show strong FR dominance with emerging DE presence. FR maintains overwhelming prevalence: 99.2\% (1999–2004), 98.7\% (2005–2009), 99.7\% (2010–2014), 99.2\% (2015–2019), and 97.3\% (2020–2025). DE shows modest but consistent growth from 0.8\% to 2.7\% (+1.9 percentage points), while EN contributions remain negligible (0.0\% across all periods). This distribution reflects the systematic morphological adaptation of FR lexical material into LU, with DE playing a secondary but growing role.

The absolute volume of morphological borrowings has grown substantially, but from a small base. Between the first (1999–2004) and most recent (2020–2025) periods, total borrowing instances increased by a factor of 13.2×, from 483 to 6{,}398. This growth pattern parallels the broader code-switching trends and reflects the expanding application of morphological adaptation rules over time, particularly for FR lexical material.

Year-over-year data confirm these trends, with code-switching rates climbing steadily: 5.19\% in 1999, 7.04\% by 2010, and 8.10\% in 2020, stabilizing around 7.88\% by 2024. DE's morphological borrowing share exhibits parallel growth from 0.6\% in 1999 to 2.7\% in 2024. These year-specific milestones reinforce the broader period-based observations and underscore the long-term nature of these linguistic developments.


Our 27-year analysis of 259,305 RTL articles shows that LU remains the matrix language while multilingual practices are pervasive but low-intensity. At the document level (RQ2), 77.1\% of articles mix LU with at least one of FR/DE/EN, yet median CMI rises only modestly with breadth (LU+1→LU+3: 3.90 → 7.00), indicating localized insertions rather than balanced bilingual text. This profile—broad but shallow mixing with a stable matrix language—is consistent with typological accounts that separate entrenched borrowing from alternational code-switching and emphasize structural dominance of the recipient language in edited prose \citep{bullock2009cambridge,poplack2017borrowing}. We summarize the supporting evidence per RQ1–RQ4:
\begin{itemize}
    \item 

    RQ1. Domain comparisons reveal small-sample spikes: the highest CMI/entropy/M-index values (CSV Süden, PK Juncker, DP Süden 2009) stem from $n\leq 3$ items and should be read as event-driven outliers, consistent with prior findings that entropy and mutual information measures can be inflated or unstable in small datasets~\cite{schroeder2004alternative}. In contrast, large sections, National (52K, CMI = 8.82) and International (45K, CMI = 8.00), exhibit stable, moderate mixing across decades, aligning with research showing more reliable entropy metrics in large datasets~\cite{hendrix2020role}.

\item RQ2. Widespread but low-intensity multilingual mixing at the document level. Despite 77\% of documents containing multiple languages, LU consistently functions as the matrix language. This pattern of broad but shallow mixing aligns with findings that dominant local languages tend to retain structural control even in heavily multilingual media environments~\cite{lam2020inter,stell2012code,bullock2009cambridge}.

\item RQ3. Token-level adaptation shows a mixed orthographic–morphological profile. From 25,444 borrowing instances, morphological changes are the collective majority (63.8\%), orthographic accounts for 35.9\%, and lexical accounts are rare (0.3\%). The largest single rules are orthographic \textit{on→oun} (34.5\%) and \textit{eur→er} (29.6\%), with productive but smaller morphological rules such as \textit{é→éiert} (20.8\%) and \textit{er→éieren} (10.4\%). This supports a Zipf-like concentration, a few highly productive rules dominate the head, while lexical-only borrowing forms a minimal tail \citep{poplack2017borrowing,riionheimo2002borrow}.

\item RQ4. Diachronically, code-switching intensifies ($\approx$ +37\% from 1999–2004 to 2020–2025), with increasingly stable period-level variation. Within morphological borrowing specifically, the overall share is small ($\approx$ 0.73\%) but grows in absolute counts (13.2×, 1999–2004 → 2020–2025). FR overwhelmingly supplies morphologically adapted items ($\approx$ 97–99\% by period), while DE shows modest growth (to 2.7\%), and EN remains negligible. These trends indicate persistent insertional code-switching as the main driver of multilingual mixing, with selective morphological assimilation concentrated in a small set of highly productive rules~\cite{grimstad2017code}.

\end{itemize}

For borrowing-aware NLP, sentence-level LID plus token-level, rule/lexicon-guided resolution reduces ambiguity where integrated forms resemble LU natives \citep{bullock2009cambridge,poplack2017borrowing}. 
For borrowing-aware NLP, sentence-level LID plus token-level, rule/lexicon-guided resolution reduces ambiguity where integrated forms resemble LU natives~\cite{poplack2017borrowing,bullock2009cambridge}, and remains critical given the limitations of sequence-based LID in code-switched environments~\cite{burchell2024code}. Given the head-heavy rule distribution and class imbalance, compact, constraint-aware decoders remain attractive, while document-level indices (CMI, entropy, M-index) are best used as context. The Simple View of Borrowing further reinforces the utility of lexical/formulaicity-based classification and low-complexity lexicon integration~\cite{treffers2025simple}, while mutual information scores and morphosyntactic listedness offer compact proxies for borrowing likelihood in ambiguous units. Reporting borrowed token/type rates, donor entropy over borrowed items, and assimilation ratios provides more actionable diagnostics for normalization and lexicon management \citep{alvarez2022detecting,poplack2017borrowing}. Similar metrics have been shown effective in tracking lexical innovation and entropy shifts during rapid borrowing events like COVID-19 media spikes~\cite{foster2021new}, supporting their use in modern NLP diagnostics. Finally, the edited-news bias likely underestimates rare adaptations; complementary conversational/social data will help surface tail phenomena. This aligns with findings that code-switching norms vary significantly between written and spoken domains, and that conversational data are more likely to surface nonce or low-frequency borrowings~\cite{dias2017code,hickey2009code}.

\section{Conclusion \& Future Work}
Our 27-year study of LU news shows a stable matrix language with pervasive but low-intensity multilingual practice: 77\% of articles mix LU with FR/DE/EN, yet median CMI remains modest even for LU+3 documents. Domain extremes are driven by small samples, whereas large sections (e.g., \textit{National}, \textit{International}) display moderate, persistent mixing. Token-level analysis reveals a head-heavy adaptation profile: borrowing instances are dominated collectively by morphological rules, alongside strong orthographic contributors (e.g., \textit{on→oun}, \textit{eur→er}); fully lexical, unadapted items are rare. Diachronically, overall code-switching intensifies, while morphologically adapted borrowings remain a small share and are overwhelmingly FR-sourced, with a modest rise in DE. Methodologically, sentence-level LID coupled with token-level, rule/lexicon-guided resolution offers practical robustness. Borrowing-first diagnostics (borrowed token/type rates, donor entropy over borrowed items, assimilation ratios) are more actionable than headline document-level indices alone for tasks such as lexicon enrichment, normalization of entrenched hybrids, and evaluation of Automatic Speech Recognition (ASR) or Machine Translation (MT) systems on loanwords.

Future work will extend LuxBorrow in four directions. First, we will create a manually annotated gold standard subset with inter-annotator agreement to better calibrate the rule-based pipeline and quantify remaining error. Second, we will add conversational and social media data to capture rarer adaptation patterns that are underrepresented in edited news. Third, we plan to combine our pattern-based detector with data-driven models, for example, sequence taggers that use the loanword registry as weak supervision, to improve recall and portability to other language pairs. Finally, we will evaluate the impact of an \texttt{is\_loan} feature in downstream tasks such as Luxembourgish normalization, NER, ASR, and machine translation.

\paragraph{Data and code availability.}
Code for preprocessing, LID, borrowing detection, and reproduction of the figures is available at
\texttt{github.com/NinaKivanani/LuxBorrow-LREC}.
Due to copyright and database rights of RTL Luxembourg, the full article texts are not redistributed, but access can be requested via
\texttt{ai@rtl.lu}; we release scripts, schemas, pattern lists, and aggregate outputs.

\section*{Limitations}
\label{sec:limitations}
Our pipeline is largely automated and not anchored in a fully manually annotated gold standard corpus. This improves scale and reproducibility but introduces uncertainty. Sentence gating and token routing depend on LID confidence, so short or ambiguous sentences can misclassify borrowing vs.\ code switching. We plan to mitigate these limits via targeted human annotation with inter annotator agreement, active learning driven expansion of patterns and lexicon, cross genre validation (speech and social media), and mixed effects modelling with domain and time controls.

LuxBorrow reflects the editorial choices, style, and topic mix of a single major media outlet, RTL Luxembourg, over 27 years. As such, it is not a representative sample of all Luxembourgish language use. Models or tools trained on this corpus may inherit RTL specific biases in topic coverage, register, and language choices. We explicitly document domain distributions and mixing patterns and encourage users of LuxBorrow to take these biases into account when drawing sociolinguistic conclusions or building downstream NLP systems.

\section{Ethical Considerations}
\label{sec:ethics}

\paragraph{Data source and legal basis}
The LuxBorrow corpus is built from online news articles published by RTL Luxembourg between 1996 and 2025. The data were obtained under a formal research collaboration between RTL Luxembourg and the University of Luxembourg, with controlled access to the internal news archive provided by an RTL collaborator. No user accounts were accessed, no technical protection measures were bypassed, and we did not perform large-scale scraping of the public website. All processing follows the applicable terms of use agreed with RTL Luxembourg and EU text and data mining provisions for non-commercial scientific research.

\paragraph{Copyright and database rights}
All articles remain under the copyright and database rights of RTL Luxembourg. Our preprocessing, annotation, and analysis operate on copies stored on secure institutional infrastructure, including the MeluXina high performance computing system, a EuroHPC supercomputer hosted and operated by LuxProvide in Luxembourg, which we access via a research allocation from the University of Luxembourg, for non commercial scientific research purposes. In line with RTL’s rights and applicable EU copyright and database rights legislation, we do not redistribute the full text of the corpus. Instead, we release only derived artefacts that do not substitute for the original content, such as
(i) the borrowing annotation schema and pattern lists,
(ii) scripts to reproduce our pipeline on any legally obtained Luxembourgish news corpus,
(iii) aggregate statistics and plots reported in this paper, and
(iv) small, non substitutable excerpted examples where required for illustration.
Researchers who require access to the underlying RTL corpus may request it via our RTL collaborator, subject to RTL’s approval and any necessary data sharing agreements.

\paragraph{Data protection and privacy}
News articles may mention identifiable individuals. These mentions concern public journalistic content that is already lawfully available online, but they remain sensitive textual data. We do not link the corpus to external personal records, profile individual persons, or attempt to infer sensitive attributes. All analyses are carried out at the level of tokens, sentences, and aggregated document sets, not at the level of individuals. The work therefore adds minimal incremental risk to the persons mentioned in the articles. Data are stored and processed on secure institutional infrastructure, including the MeluXina high performance computing system hosted and operated by LuxProvide in Luxembourg, in compliance with the GDPR and applicable national data protection requirements.

\paragraph{Normative use and impact}
As discussed in Section~\ref{sec:limitations}, LuxBorrow reflects the editorial choices and topic mix of a single media outlet and is not a representative sample of all Luxembourgish language use. Our borrowing first diagnostics are intended as a descriptive tool for contact linguistics and multilingual NLP, not as a prescriptive standard for “correct” Luxembourgish. We discourage the use of this resource for normative policing of lexical borrowing or for applications that marginalise code switching practices in everyday multilingual communication. Aggregate patterns of borrowing or code mixing may correlate with social or regional factors, but such correlations should be interpreted with caution to avoid reifying stereotypes or overgeneralising from a single media source.


\section{Acknowledgements}
We thank RTL Luxembourg and Tom Weber for providing access to the news archive and for supporting the use of these data for research. This work was carried out within the project ``Comprehensive Text-to-Speech Development for Luxembourgish with Emotional Enhancements (LuxVoice),'' project reference 19205922. The present project was supported by the National Research Fund, Luxembourg.

\section{Bibliographical References}\label{sec:reference}

\bibliographystyle{lrec2026-natbib}
\bibliography{lrec2026-example}

\begin{thebibliography}{6}
\expandafter\ifx\csname natexlab\endcsname\relax\def\natexlab#1{#1}\fi

\bibitem[{{Wiktionary}(2025{\natexlab{a}})}]{Wiktionary_EN_from_FR}
{Wiktionary}. 2025{\natexlab{a}}.
\newblock \emph{Category:English terms borrowed from French – Wiktionary}.
\newblock [Accessed on 20-10-2025]. Available at: \url{https://en.wiktionary.org/wiki/Category:English_terms_borrowed_from_French}.

\bibitem[{{Wiktionary}(2025{\natexlab{b}})}]{Wiktionary_Old_High_German}
{Wiktionary}. 2025{\natexlab{b}}.
\newblock \emph{Category:German terms derived from Old High German – Wiktionary}.
\newblock [Accessed on 20-10-2025]. Available at: \url{https://en.wiktionary.org/wiki/Category:German_terms_derived_from_Old_High_German}.

\bibitem[{{Wiktionary}(2025{\natexlab{c}})}]{Wiktionary_FR_from_EN}
{Wiktionary}. 2025{\natexlab{c}}.
\newblock \emph{Catégorie:Anglicismes en français – Wiktionary}.
\newblock [Accessed on 20-10-2025]. Available at: \url{https://fr.wiktionary.org/wiki/Cat%C3%A9gorie:Anglicismes_en_fran%C3%A7ais}.

\bibitem[{{Wiktionary}(2025{\natexlab{d}})}]{Wiktionary_DE_from_EN}
{Wiktionary}. 2025{\natexlab{d}}.
\newblock \emph{Verzeichnis:Deutsch/Anglizismen – Wiktionary}.
\newblock [Accessed on 25-09-2025]. Available at: \url{https://de.wiktionary.org/wiki/Verzeichnis:Deutsch/Anglizismen}.

\bibitem[{{Wiktionary}(2025{\natexlab{e}})}]{Wiktionary_DE_from_FR}
{Wiktionary}. 2025{\natexlab{e}}.
\newblock \emph{Verzeichnis:Deutsch/Gallizismen – Wiktionary}.
\newblock [Accessed on 25-09-2025]. Available at: \url{https://de.wiktionary.org/wiki/Verzeichnis:Deutsch/Gallizismen}.

\bibitem[{{Zenter fir d'Lëtzebuerger Sprooch}(2025)}]{LOD}
{Zenter fir d'Lëtzebuerger Sprooch}. 2025.
\newblock \href {https://lod.lu} {\emph{Lëtzebuerger Online Dictionnaire (LOD)}}.
\newblock Official reference dictionary for Luxembourgish.

\end{thebibliography}


\begin{thebibliography}{43}
\expandafter\ifx\csname natexlab\endcsname\relax\def\natexlab#1{#1}\fi

\bibitem[{Aguilar et~al.(2020)Aguilar, Kar, and Solorio}]{aguilar2020lince}
Gustavo Aguilar, Sudipta Kar, and Thamar Solorio. 2020.
\newblock Lince: A centralized benchmark for linguistic code-switching evaluation.
\newblock In \emph{Proceedings of the Twelfth Language Resources and Evaluation Conference}, pages 1803--1813.

\bibitem[{{\'A}lvarez-Mellado and Lignos(2022)}]{alvarez2022detecting}
Elena {\'A}lvarez-Mellado and Constantine Lignos. 2022.
\newblock Detecting unassimilated borrowings in spanish: An annotated corpus and approaches to modeling.
\newblock In \emph{Proceedings of the 60th Annual Meeting of the Association for Computational Linguistics (Volume 1: Long Papers)}, pages 3868--3888.

\bibitem[{Bullock and Toribio(2009)}]{bullock2009cambridge}
Barbara~E Bullock and Almeida~Jacqueline Toribio. 2009.
\newblock \emph{The Cambridge handbook of linguistic code-switching}.
\newblock Cambridge university press.

\bibitem[{Burchell et~al.(2023)Burchell, Birch, Bogoychev, and Heafield}]{burchell-etal-2023-open}
Laurie Burchell, Alexandra Birch, Nikolay Bogoychev, and Kenneth Heafield. 2023.
\newblock \href {https://doi.org/10.18653/v1/2023.acl-short.75} {An open dataset and model for language identification}.
\newblock In \emph{Proceedings of the 61st Annual Meeting of the Association for Computational Linguistics (Volume 2: Short Papers)}, pages 865--879, Toronto, Canada. Association for Computational Linguistics.

\bibitem[{Burchell et~al.(2024)Burchell, Birch, Thompson, and Heafield}]{burchell2024code}
Laurie Burchell, Alexandra Birch, Robert Thompson, and Kenneth Heafield. 2024.
\newblock Code-switched language identification is harder than you think.
\newblock In \emph{Proceedings of the 18th Conference of the European Chapter of the Association for Computational Linguistics (Volume 1: Long Papers)}, pages 646--658.

\bibitem[{Chan(2025)}]{chan2025borrowing}
Brian Hok-Shing Chan. 2025.
\newblock Borrowing or code-switching? single-word english prepositions in hong kong cantonese.
\newblock \emph{Open Linguistics}, 11(1):20250045.

\bibitem[{Chandu et~al.(2018)Chandu, Loginova, Gupta, van Genabith, Neumann, Chinnakotla, Nyberg, and Black}]{chandu2018code}
Khyathi Chandu, Ekaterina Loginova, Vishal Gupta, Josef van Genabith, G{\"u}nter Neumann, Manoj Chinnakotla, Eric Nyberg, and Alan~W Black. 2018.
\newblock Code-mixed question answering challenge: Crowd-sourcing data and techniques.
\newblock In \emph{Proceedings of the Third Workshop on Computational Approaches to Linguistic Code-Switching}, pages 29--38.

\bibitem[{de~la Rosa(2021)}]{de2021adobo}
Javier de~la Rosa. 2021.
\newblock Adobo 2021: The futility of stilts for the classification of lexical borrowings in spanish.
\newblock In \emph{IberLEF@ SEPLN}, pages 947--955.

\bibitem[{Deuchar(2020)}]{deuchar2020code}
Margaret Deuchar. 2020.
\newblock Code-switching in linguistics: A position paper.
\newblock \emph{Languages}, 5(2):22.

\bibitem[{Dias(2017)}]{dias2017code}
Silvia~Fortuna Dias. 2017.
\newblock \emph{Code-Switching and Lexical Borrowing among Brazilian Portuguese and English Bilinguals in Chicagoland}.
\newblock Ph.D. thesis, University of Illinois Chicago.

\bibitem[{Foster and Welsh(2021)}]{foster2021new}
Stuart~Mannix Foster and Alistair Welsh. 2021.
\newblock A ‘new normal’of code-switching: Covid-19, the indonesian media and language change.
\newblock \emph{Indonesian Journal of Applied Linguistics}, 11:1.

\bibitem[{Grimstad(2017)}]{grimstad2017code}
Maren~Berg Grimstad. 2017.
\newblock The code-switching/borrowing debate: evidence from english-origin verbs in american norwegian.
\newblock \emph{Lingue e linguaggio}, 16(1):3--34.

\bibitem[{Hendrix and Sun(2020)}]{hendrix2020role}
Peter Hendrix and Ching~Chu Sun. 2020.
\newblock The role of information theory for compound words in mandarin chinese and english.
\newblock \emph{Cognition}, 205:104389.

\bibitem[{Hickey(2009)}]{hickey2009code}
Tina Hickey. 2009.
\newblock Code-switching and borrowing in irish 1.
\newblock \emph{Journal of Sociolinguistics}, 13(5):670--688.

\bibitem[{Kent and Claeser(2018)}]{kent2018incorporating}
Samantha Kent and Daniel Claeser. 2018.
\newblock Incorporating code-switching and borrowing in dutch-english automatic language detection on twitter.
\newblock In \emph{Proceedings of the Future Technologies Conference}, pages 418--434. Springer.

\bibitem[{Khanuja et~al.(2020)Khanuja, Dandapat, Srinivasan, Sitaram, and Choudhury}]{khanuja2020gluecos}
Simran Khanuja, Sandipan Dandapat, Anirudh Srinivasan, Sunayana Sitaram, and Monojit Choudhury. 2020.
\newblock Gluecos: An evaluation benchmark for code-switched nlp.
\newblock In \emph{Proceedings of the 58th Annual Meeting of the Association for Computational Linguistics}, pages 3575--3585.

\bibitem[{Lam and Matthews(2020)}]{lam2020inter}
Chit~Fung Lam and Stephen Matthews. 2020.
\newblock Inter-sentential code-switching and language dominance in cantonese--english bilingual children.
\newblock \emph{Journal of Monolingual and Bilingual Speech}, 2(1):73--105.

\bibitem[{List and Forkel(2022)}]{list2022automated}
Johann-Mattis List and Robert Forkel. 2022.
\newblock Automated identification of borrowings in multilingual wordlists.
\newblock \emph{Open Research Europe}, 1:79.

\bibitem[{Lutgen et~al.(2025)Lutgen, Plum, Purschke, and Plank}]{lutgen2025neural}
Anne-Marie Lutgen, Alistair Plum, Christoph Purschke, and Barbara Plank. 2025.
\newblock Neural text normalization for luxembourgish using real-life variation data.
\newblock In \emph{Proceedings of the 12th Workshop on NLP for Similar Languages, Varieties and Dialects}, pages 115--127.

\bibitem[{Mager et~al.(2019)Mager, {\c{C}}etino{\u{g}}lu, and von~der Wense}]{mager2019subword}
Manuel Mager, {\"O}zlem {\c{C}}etino{\u{g}}lu, and Katharina von~der Wense. 2019.
\newblock Subword-level language identification for intra-word code-switching.
\newblock In \emph{Proceedings of the 2019 Conference of the North American Chapter of the Association for Computational Linguistics: Human Language Technologies, Volume 1 (Long and Short Papers)}, pages 2005--2011.

\bibitem[{Malon and Zhu(2024)}]{malon2024self}
Christopher Malon and Xiaodan Zhu. 2024.
\newblock Self-consistent decoding for more factual open responses.
\newblock \emph{arXiv preprint arXiv:2403.00696}.

\bibitem[{Masoj{\'c}(2023)}]{masojc2023borrowings}
Irena Masoj{\'c}. 2023.
\newblock Borrowings or code-switching? analysis of ways of morphosyntactic integration of lithuanian and russian nouns in the novel" robczik" by bartosz po{\l}o{\'n}ski.
\newblock \emph{Slavistica Vilnensis}, 68(1):87--101.

\bibitem[{Mellado and Lignos(2022)}]{mellado2022borrowing}
Elena~{\'A}lvarez Mellado and Constantine Lignos. 2022.
\newblock Borrowing or codeswitching? annotating for finer-grained distinctions in language mixing.
\newblock In \emph{Proceedings of the Thirteenth Language Resources and Evaluation Conference}, pages 3195--3201.

\bibitem[{Mi et~al.(2020)Mi, Xie, and Zhang}]{mi2020loanword}
Chenggang Mi, Lei Xie, and Yanning Zhang. 2020.
\newblock Loanword identification in low-resource languages with minimal supervision.
\newblock \emph{ACM Transactions on Asian and Low-Resource Language Information Processing (TALLIP)}, 19(3):1--22.

\bibitem[{Miller(2021)}]{miller2021neural}
John~Edward Miller. 2021.
\newblock Neural borrowing detection with monolingual lexical models.
\newblock In \emph{Proceedings of the student research workshop associated with RANLP 2021}.

\bibitem[{Plum et~al.(2024)Plum, D{\"o}hmer, Milano, Lutgen, and Purschke}]{plum2024luxbank}
Alistair Plum, Caroline D{\"o}hmer, Emilia Milano, Anne-Marie Lutgen, and Christoph Purschke. 2024.
\newblock Luxbank: The first universal dependency treebank for luxembourgish.
\newblock \emph{TLT 2024}, page~30.

\bibitem[{Plum et~al.(2025)Plum, Ranasinghe, and Purschke}]{plum2025text}
Alistair Plum, Tharindu Ranasinghe, and Christoph Purschke. 2025.
\newblock Text generation models for luxembourgish with limited data: A balanced multilingual strategy.
\newblock In \emph{Proceedings of the 12th Workshop on NLP for Similar Languages, Varieties and Dialects}, pages 93--104.

\bibitem[{Poplack(2017)}]{poplack2017borrowing}
Shana Poplack. 2017.
\newblock \emph{Borrowing: Loanwords in the Speech Community and in the Grammar}.
\newblock Oxford University Press.

\bibitem[{Poplack and Sankoff(1984)}]{poplack1984borrowing}
Shana Poplack and David Sankoff. 1984.
\newblock Borrowing: the synchrony of integration.

\bibitem[{Poplack et~al.(1988)Poplack, Sankoff, and Miller}]{poplack1988social}
Shana Poplack, David Sankoff, and Christopher Miller. 1988.
\newblock The social correlates and linguistic processes of lexical borrowing and assimilation.
\newblock \emph{Linguistics}, 26(1):47--104.

\bibitem[{Prabhugaonkar et~al.(2017)Prabhugaonkar, Peketi, Ganeshan, and Sureshkumar}]{prabhugaonkar2017differentiating}
Neha Prabhugaonkar, Sai~Kiran Peketi, Kavita Ganeshan, and Unnikrishnan Sureshkumar. 2017.
\newblock Differentiating code-borrowing from code-mixing.
\newblock In \emph{Proceedings of the 4th ACM IKDD Conferences on Data Sciences}, pages 1--2.

\bibitem[{Purschke(2020)}]{spellux2020}
Christoph Purschke. 2020.
\newblock \href {http://arxiv.org/abs/https://orbilu.uni.lu/10993/42807} {spellux – automatic text normalization for luxembourgish}.

\bibitem[{Riionheimo(2002)}]{riionheimo2002borrow}
Helka Riionheimo. 2002.
\newblock How to borrow a bound morpheme? evaluating the status of structural interference in a contact between closely-related languages.
\newblock \emph{Finnish Journal of Linguistics}, (15):187--218.

\bibitem[{Rosillo-Rodes et~al.(2025)Rosillo-Rodes, San~Miguel, and S{\'a}nchez}]{rosillo2025entropy}
Pablo Rosillo-Rodes, Maxi San~Miguel, and David S{\'a}nchez. 2025.
\newblock Entropy and type-token ratio in gigaword corpora.
\newblock \emph{Physical Review Research}, 7(3):033054.

\bibitem[{Sabty et~al.(2021)Sabty, Mesabah, {\c{C}}etino{\u{g}}lu, and Abdennadher}]{sabty2021language}
Caroline Sabty, Islam Mesabah, {\"O}zlem {\c{C}}etino{\u{g}}lu, and Slim Abdennadher. 2021.
\newblock Language identification of intra-word code-switching for arabic--english.
\newblock \emph{Array}, 12:100104.

\bibitem[{Schroeder(2004)}]{schroeder2004alternative}
Marcin~J Schroeder. 2004.
\newblock An alternative to entropy in the measurement of information.
\newblock \emph{Entropy}, 6(5):388--412.

\bibitem[{Soto and Hirschberg(2018)}]{soto2018joint}
Victor Soto and Julia Hirschberg. 2018.
\newblock Joint part-of-speech and language id tagging for code-switched data.
\newblock In \emph{Proceedings of the Third Workshop on Computational Approaches to Linguistic Code-Switching}, pages 1--10.

\bibitem[{Srivastava and Singh(2021)}]{srivastava2021challenges}
Vivek Srivastava and Mayank Singh. 2021.
\newblock Challenges and limitations with the metrics measuring the complexity of code-mixed text.
\newblock In \emph{Proceedings of the Fifth Workshop on Computational Approaches to Linguistic Code-Switching}, pages 6--14.

\bibitem[{Stell and Couto(2012)}]{stell2012code}
Gerald Stell and Maria Del Carmen~Parafita Couto. 2012.
\newblock Code-switching practices in luxembourg’s lusophone minority: a pilot study on how an immigrant community linguistically behaves differently from the majority.
\newblock \emph{Zeitschrift fur Sprachwissenschaft}, 31(1):153--185.

\bibitem[{Thara and Poornachandran(2018)}]{thara2018code}
S~Thara and Prabaharan Poornachandran. 2018.
\newblock Code-mixing: A brief survey.
\newblock In \emph{2018 International conference on advances in computing, communications and informatics (ICACCI)}, pages 2382--2388. IEEE.

\bibitem[{Treffers-Daller(2023)}]{treffers2025simple}
Jeanine Treffers-Daller. 2023.
\newblock The simple view of borrowing and code-switching.
\newblock \emph{International Journal of Bilingualism}, 29(2):347--370.

\bibitem[{Winata et~al.(2023)Winata, Aji, Yong, and Solorio}]{winata2023decades}
Genta~Indra Winata, Alham~Fikri Aji, Zheng-Xin Yong, and Thamar Solorio. 2023.
\newblock The decades progress on code-switching research in nlp: A systematic survey on trends and challenges.
\newblock In \emph{Findings of the Association for Computational Linguistics: ACL 2023}, pages 2936--2978.

\bibitem[{Zha et~al.(2023)Zha, Yang, Li, and Hu}]{zha2023alignscore}
Yuheng Zha, Yichi Yang, Ruichen Li, and Zhiting Hu. 2023.
\newblock Alignscore: Evaluating factual consistency with a unified alignment function.
\newblock In \emph{Proceedings of the 61st Annual Meeting of the Association for Computational Linguistics (Volume 1: Long Papers)}, pages 11328--11348.

\end{thebibliography}

\section{Language Resource References}

\label{lr:ref}
\bibliographystylelanguageresource{lrec2026-natbib}
\bibliographylanguageresource{languageresource}

\section{Appendices}

\subsection{Loanword Identification}
Table \ref{tab:pattern_examples} summarizes the selected transformation patterns used to identify loanwords.

\begin{table}[hpt!]
    \centering
    \small
    \begin{tabular}{|c|c|c|}
\hline
\textbf{Pattern} & \textbf{LU Word} & \textbf{Source Word} \\
\hline
\multirow{3}{*}{nner$\to$néieren}
  & sanctionéieren & sanctionner \\
  & ordonéieren & ordonner \\
  & deconéieren & deconner \\
\hline
\multirow{3}{*}{exact}
  & Vignette & vignette \\
  & sektoriell & sektoriell \\
  & Absence & absence \\
\hline
\multirow{3}{*}{er$\to$éieren}
  & filetéieren & fileter \\
  & immuniséieren & immuniser \\
  & marginaliséieren & marginaliser \\
\hline
\multirow{3}{*}{le$\to$el}
  & Decapotabel & décapotable \\
  & accessibel & accessible \\
  & favorabel & favorable \\
\hline
\multirow{3}{*}{on$\to$oun}
  & Motivatioun & motivation \\
  & Abstentioun & abstention \\
  & Exploitatioun & exploitation \\
\hline
\multirow{3}{*}{-e}
  & Karwoch & Karwoche \\
  & Schlei & Schleie \\
  & Manikür & Maniküre \\
\hline
\multirow{3}{*}{é$\to$éit}
  & Stabilitéit & stabilité \\
  & Majestéit & Majesté \\
  & Disponibilitéit & disponibilité \\
\hline
\multirow{3}{*}{é$\to$éiert}
  & bornéiert & borné \\
  & arméiert & armé \\
  & accidentéiert & accidenté \\
\hline
\multirow{3}{*}{\parbox[c]{2cm}{\centering on$\to$oun \\ + \\ c$\to$k}}
  & Fiktioun & fiction \\
  & Kollektioun & collection \\
  & Attraktioun & attraction \\
\hline
\multirow{3}{*}{ir$\to$éieren}
  & reusséieren & reussir \\
  & investéieren & investir \\
  & demoléieren & demolir \\
\hline
\multirow{3}{*}{ät$\to$éit}
  & Skurrilitéit & Skurrilität \\
  & Aggressivitéit & Aggressivität \\
  & Fakultéit & Fakultät \\
\hline
\multirow{3}{*}{que$\to$ck}
  & Barock & baroque \\
  & Attack & attaque \\
  & Pick & pique \\
\hline
\multirow{3}{*}{eur$\to$er}
  & Rapporter & rapporteur \\
  & Coiffer & coiffeur \\
  & Entreprener & entrepreneur \\
\hline
\multirow{3}{*}{É$\to$E}
  & Eclat & éclat \\
  & Epicière & épicière \\
  & Ecart & écart \\
\hline
\multirow{2}{*}{i$\to$éiert}
  & introvertéiert & introverti \\
  & etabléiert & etabli \\
\hline
\multirow{2}{*}{nné$\to$néiert}
  & proportionéiert & proportionné \\
  & passionéiert & passionné \\
\hline
\end{tabular}
    \caption{Examples of orthographic and morphological transformations in LU loanwords.}
    \label{tab:pattern_examples}
\end{table}

\end{document}